\documentclass{article}

\usepackage[preprint]{neurips_2025}

\usepackage[utf8]{inputenc} 
\usepackage[T1]{fontenc}    
\usepackage{hyperref}       
\usepackage{url}            
\usepackage{booktabs}       
\usepackage{amsfonts}       
\usepackage{graphicx}       
\usepackage{nicefrac}       
\usepackage{microtype}      
\usepackage{xcolor}         
\usepackage{booktabs}
\usepackage{amsmath}

\usepackage{tabularx}
\usepackage{float}
\usepackage{booktabs}
\usepackage{siunitx} 
\usepackage[most]{tcolorbox}  
\usepackage{fancyvrb}         
\usepackage{fvextra} 
\usepackage{enumitem}                 

\definecolor{violet}{RGB}{138,43,226}

\title{LitBench: A Benchmark and Dataset for Reliable Evaluation of Creative Writing}


\author{\bfseries
    Daniel Fein\footnotemark[1] \quad
  Sebastian Russo\footnotemark[1] \quad
  Violet Xiang\footnotemark[1] \quad Kabir Jolly \\[2pt]
    \bfseries
    Rafael Rafailov \quad Nick Haber \\[4pt]
    \mdseries
    Stanford University
}

\begin{document}
\maketitle
\footnotetext[1]{Equal contribution. Correspondence: \texttt{drfein@stanford.edu}.}
\begin{abstract}
Evaluating creative writing generated by large language models (LLMs) remains challenging because open-ended narratives lack ground truths. Without performant automated  evaluation methods, off-the-shelf (OTS) language models are employed as zero-shot judges, yet their reliability is unclear in this context. In pursuit of robust evaluation for creative writing, we introduce LitBench, the first standardized benchmark and paired dataset for creative writing verification, comprising a held-out test set of 2,480 debiased, human-labeled story comparisons drawn from Reddit and a 43,827-pair training corpus of human preference labels. Using LitBench, we (i) benchmark zero‐shot LLM judges, (ii) train Bradley–Terry and generative reward models, and (iii) conduct an online human study to validate reward model rankings on newly LLM-generated stories. Our benchmark identifies Claude‑3.7‑Sonnet as the strongest off‑the‑shelf judge, reaching 73\% agreement with human preferences; among trained reward models, Bradley-Terry and Generative reward models both attain an accuracy of 78\%, outperforming all off‑the‑shelf judges. An online human study further confirms that our trained reward models consistently align with human preferences in novel LLM-generated stories. We release LitBench and reward models \href{https://huggingface.co/collections/SAA-Lab/litbench-68267b5da3aafe58f9e43461}{here}, providing a vetted resource for reliable, automated evaluation and optimization of creative‑writing systems.
\end{abstract}

\label{sec:intro}
\section{Introduction}
Automated verification with oracles or learned verifiers has catalyzed rapid progress in math and code generation \citep{hendrycks2021measuring, gao2024omni, jimenez2023swe, pan2024training}. By contrast, creative writing is inherently divergent: given the same prompt, authors may produce different yet equally valid stories. The lack of ground truth labels hinders verification and, consequently, progress in creative writing generation. Evaluation by human experts with structured rubric is reliable, but it is expensive to collect such judgements, particularly at the scale of AI–generated text \citep{chakrabarty2024art}. In domains where human where ground truth are usually collected from human raters, LLM judges are often used (\citep{badshah2024referenceguidedverdictllmsasjudgesautomatic}; \citep{son2024llmasajudgerewardmodel}). 

The agreement between LLM judgments and human preferences has been found to be reasonable in the contexts of dialog, helpfulness, and summarization tasks \citep{zheng2023judging}. But, they exhibit biases, such as favoring lengthy text \citep{wang2023largelanguagemodelsfair}, and lack of internal consistency \citep{wei2025systematicevaluationllmasajudgellm}. \cite{feuer2025styleoutweighssubstancefailure} found that within these tasks, stylistic choices account for the judgments of language models more often than substance. This raises questions about the reliability of judges in the context of creative writing, where form and content are paramount. 

We introduce \textbf{LitBench}, the first standardized benchmark of high-quality, pairwise creative writing samples, derived from Reddit's \texttt{r/WritingPrompts}. LitBench is designed to both evaluate existing zero-shot judges and enable the development of learned verifiers that better align with human preferences. Then, to study the gap between LLM-judges and trained reward models, we curate a dataset of 43k further pairwise examples from \texttt{r/WritingPrompts}. LitBench evaluation reveals that small and open-source LLM-judges fail to evaluate creative writing accurately, but that some leading proprietary models are competitive with trained verifiers. Our investigation of various reward models reveals that generative reward models (GenRMs) are on-par with Bradley-Terry reward models in this domain. While \citep{mahan2024generative} found that training GenRMs with chain-of-thought (CoT) can lead to similar or even improved performance on some preference based benchmarks, such as RewardBench, we find it hinders performance in creative writing verification, even when CoTs are distilled from a much stronger out-of-the-box verifier model. Additional human evaluation on LLM-generated stories validates that well-performing reward models on our benchmark can indeed judge creative quality. 

Our contributions are as follows.
\begin{itemize}
    \item A benchmark of 2.5k pairwise comparisons of human-written stories, coupled with a filtered and labeled training dataset for verifiers consisting of 43k pairwise examples, along with generated rationales. 
    \item Benchmarking of current approaches of creative writing verification, revealing that the best zero-shot LLM judge (\texttt{Claude-Sonnet-3.7}) underperformed small reward models (1B-7B) trained on our training set, suggesting we can get higher quality reward models at lower cost.
    \item Study of GenRMs showing that distilled chain-of-thought degrades performance for creative writing verification.
    \item Human evaluation validating that a verifier performing well on LitBench can be used to select higher-quality creative writing.
\end{itemize}

\section{Related Work}
\label{gen_inst}

\subsection{Verification}
\label{sec:verification}

 Recently, math and coding benchmarks with ground truth labels have facilitated progress in these domains \citep{gao2024omni, jimenez2023swe}. \cite{cobbe2021gsm8k} first used inference-time verification to bootstrap language model performance on GSM8K by ranking candidate solutions from a generator. More recently, \cite{costello2025thinkprunetrainimprove} and \cite{zelikman2024star} have shown that ground truth pruning of generations and retraining can improve the latent ability to correctly solve math problems. 

Without ground truth labels, verification is difficult. Reinforcement learning from human feedback (RLHF) has emerged as the dominant paradigm to align model language and behavior with human taste and steer models to follow instructions \citep{ouyang2022traininglanguagemodelsfollow, stiennon2020learning}. \cite{bai2022constitutional} developed a lower-cost method of alignment with human preferences by substituting humans for language models in the feedback process. LLM-judges have been found to agree with human preferences in some contexts \cite{zheng2023judging, liu2023geval}, though their agreement with humans in the evaluation of creative writing or other forms of artistic expression has not been systematically evaluated. 

In another attempt to avoid costly human preferences, \citep{ethayarajh2022vusable} released The Stanford Human Preferences (SHP) dataset, leveraging Reddit to distill human preferences for helpfulness using post and comment annotations. Within the genre of creative writing, \cite{chung2025modifyinglargelanguagemodel} point out that ``having a robust reward model for creative
writing is difficult due to subjectivity in evaluation.'' Existing work towards automatic story evaluation often makes use of a reference work either using a language model or metrics like BLEU and ROGUE scores \citep{li2025automatedcreativityevaluationlarge, netisopakul2023comparison}. However, \cite{fan2018hierarchical} notes that ``in our open-ended generation setting, these are not useful.'' There have been efforts for evaluation in open-ended settings, for example, in detecting narrative incoherence using structural cues, but story coherence alone does not encapsulate creative qualities which our work is interested in \citep{alihosseini2019jointly, li2020relationqualitydiversity}. 

\subsection{Creativity in Writing}
Common usage of \textit{``creativity''} lacks precision, encapsulating a multiplicity of ideas. The term suggests a harmony of integrated elements; it often implies discovery or surprise; and paradoxically, the word can describe both technical excellence and technical incompetence -- that which is fresh, unshackled, in defiance of convention \citep{barzun1960cultsofresearch}. To narrow our scope, we offer Rhode's framework for creativity: the four P's of creativity, i.e., (1) people, (2) process, (3) press, (4) products \citep{rhodes1961analysiscreativity}. This paper is concerned with creative products, in the form of short stories. Among the properties proposed for creative products, interdisciplinary theories have distilled two fundamental criteria: novelty and value \citep{callan2023howinterestingcoherentstories}. 
Our work is, in part, inspired by observing that there exist convergent human preferences for creative products. These preferences are available for display in our cultural institutions. For example, the selections of required readings in MFA program syllabi demonstrate enormous inter-institution uniformity \citep{manery2016educationcreativewritingteacher}. This extends to many creative domains, where there is a popular notion of canon works. Psychology studies have also shown convergent creative preferences, showing that expert writers achieve high agreement when asked to rank poetry and prose. \citep{amabile1982socialpsychologycreativity}. 

Judgement of written language is rooted human emotion (e.g. finding something humorous) and, by extension, in shared human experience. These abilities are not available to language models. What is available to language models, though, is aggregated human preferences for creative products. Below, we present an aggregated collection of creative \textit{products}, which we hope will further computational methods for creative \textit{process}. In other words, perhaps aligning models with human creativity amounts to curating a great reading list.

\section{LitBench}
\label{sec:dataset_curation}
LitBench is a benchmark for reward models that can judge creative writing quality, coupled with a training set which improvements can be made on. Since reward hacking is a common issue when using preference-based reward models to improve model capabilities, we carefully construct both our evaluation and training set to ensure quality. We describe the procedure in detail, and demonstrate our curation procedure is indeed helpful.
\begin{figure}[ht]
  \centering
  \includegraphics[width=1.0\linewidth]{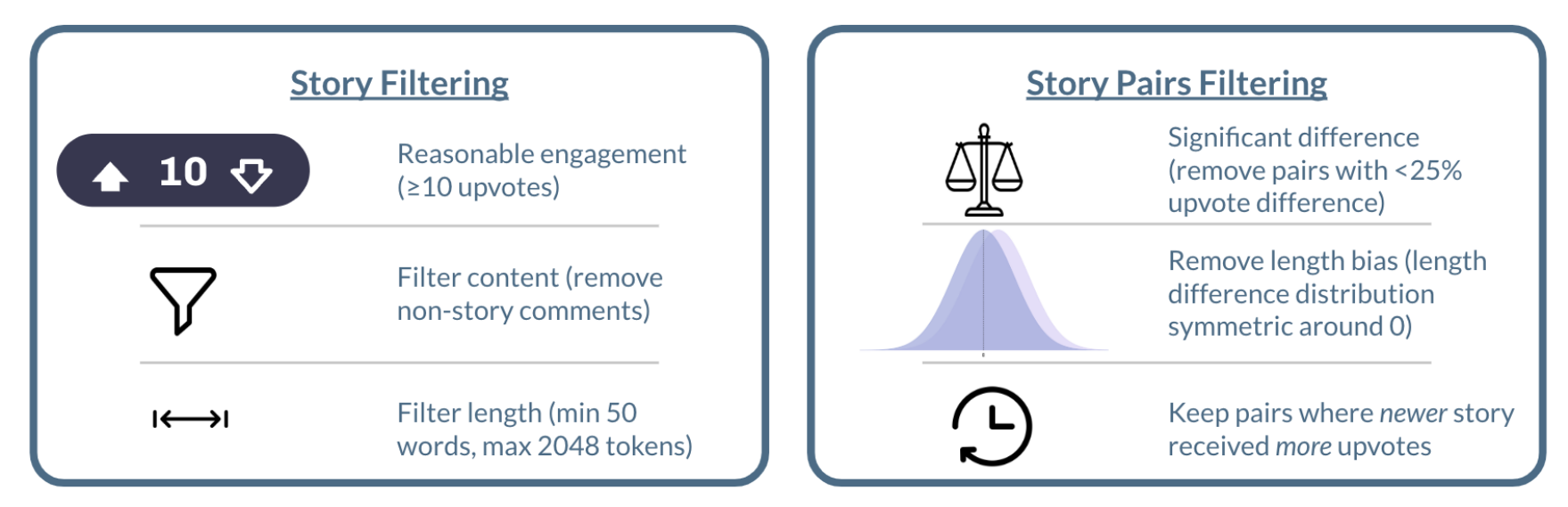}
  \caption{Preprocessing methodology for dataset creation.}
  \label{fig:data-flow}
\end{figure}

\subsection{Data Collection}
We collect writing samples from the \texttt{r/WritingPrompts} subreddit, which has 18.9 Million subscribers. Users write stories in response to writing prompts, and freely engage with posted stories through upvotes or comments. In total, \texttt{r/WritingPrompts} has amassed over one million stories. Such a large corpus enables highly selectivedata filtration, leaving data for which we can confidently assume human preferences signal. To collect the data for our benchmark, we use the Reddit API via the \texttt{praw} library. Specifically, we use the search function to collect the 100 top search results for each post collected by \citep{fan2018hierarchicalneuralstorygeneration}. This yields  5,000+ post-ids (individual prompts within the framework of the subreddit). We then construct our test set by filtering out any stories older than 2023, as this data potentially overlaps with our training dataset, and is more likely to have been included in the pretraining of the models we study here. To collect our training dataset, we curate examples from the MIT-licensed \texttt{euclaise/WritingPrompts\_preferences} dataset from Hugging Face\footnote{\href{https://huggingface.co/datasets/euclaise/WritingPrompts_preferences}{https://huggingface.co/datasets/euclaise/WritingPrompts\_preferences}}, which contains story posts from prior 2023.

\subsection{Quality Control}

We begin constructing our dataset by filtering stories independently. First, to reduce the affect of noise from small up-vote counts, we guarantee that each story has a reasonable amount of engagement by filtering out stories with fewer than 10 up-votes. Then, consistent with \citep{chung2025modifyinglargelanguagemodel}, we filter out stories with greater than 2048 tokens to remove excessively long stories. Lastly, we remove all entries with fewer than 50 words, because we find qualitatively that these are not sufficiently long to reflect the genre of creative fiction.

To form pairs, we carry out two steps to ensure that true preferences are being captured, and then one step to address length bias. Initially, we exclude pairs with marginal differences in up-votes, filtering out those with an upvote difference less than 25\%. Next, following the methodology of \citep{pmlr-v162-ethayarajh22a}, we only create pairs where the higher-upvote story is also published later, mitigating temporal bias from varying exposure durations.

\begin{figure}[htbp]
  \centering
  \includegraphics[height=0.4\linewidth]{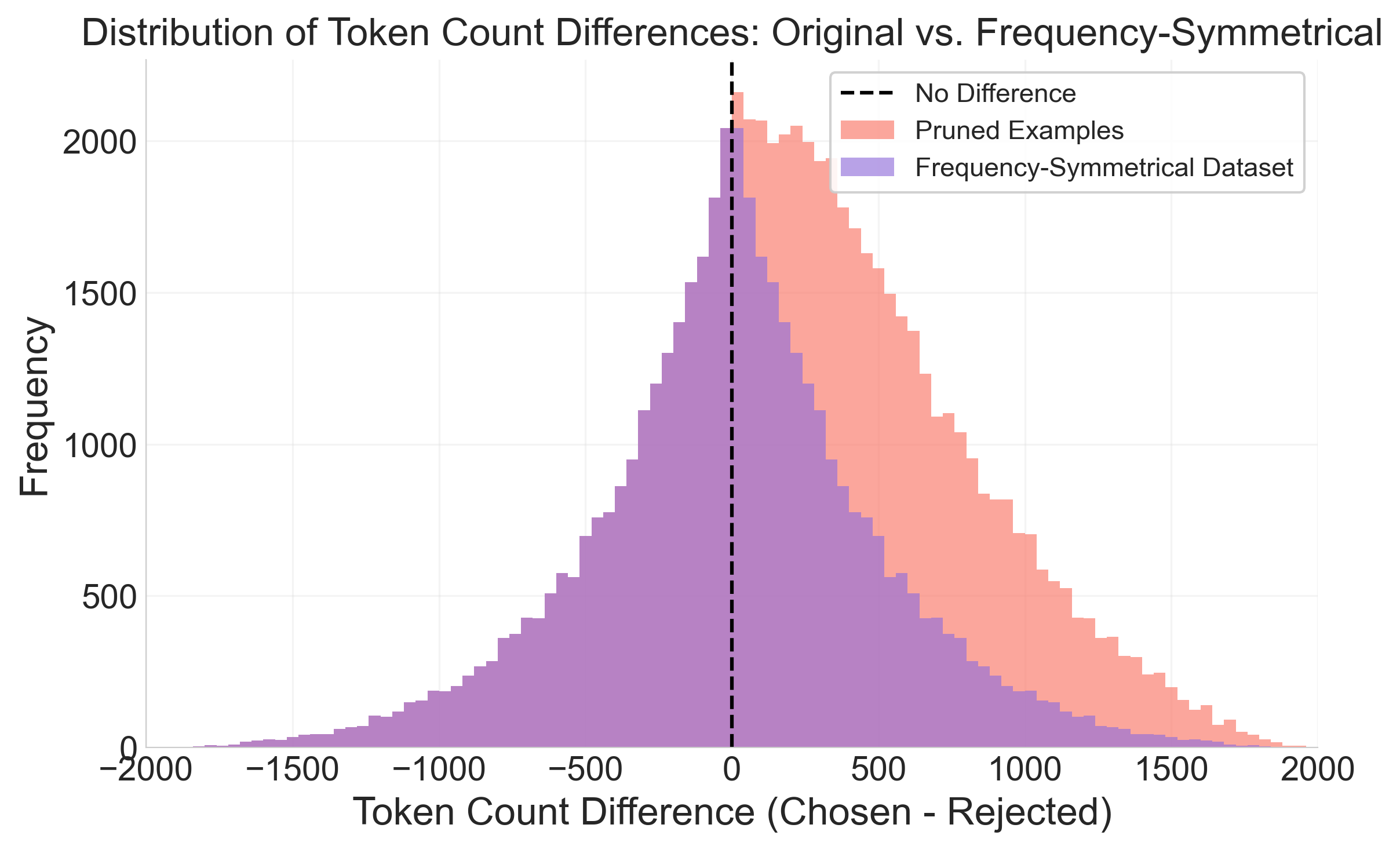}
  \caption{Length bias mitigation.}
  \label{fig:length-bias}
\end{figure}

Lastly, we find that the resultant dataset has a length bias, with $65.25\%$ of chosen responses longer than rejected responses. To address this while preserving length diversity, we construct a histogram of length differences (100 buckets) and prune pairs until achieving symmetry -- balanced proportions where chosen stories are both shorter and longer. This step is represented in Figure \ref{fig:length-bias}. The data preparation workflow is summarized in Figure \ref{fig:data-flow}.This entire process is performed independently for both our benchmark and training dataset.

\subsection{Final Dataset Description}
\begin{figure}
    \centering
    \includegraphics[width=1\linewidth]{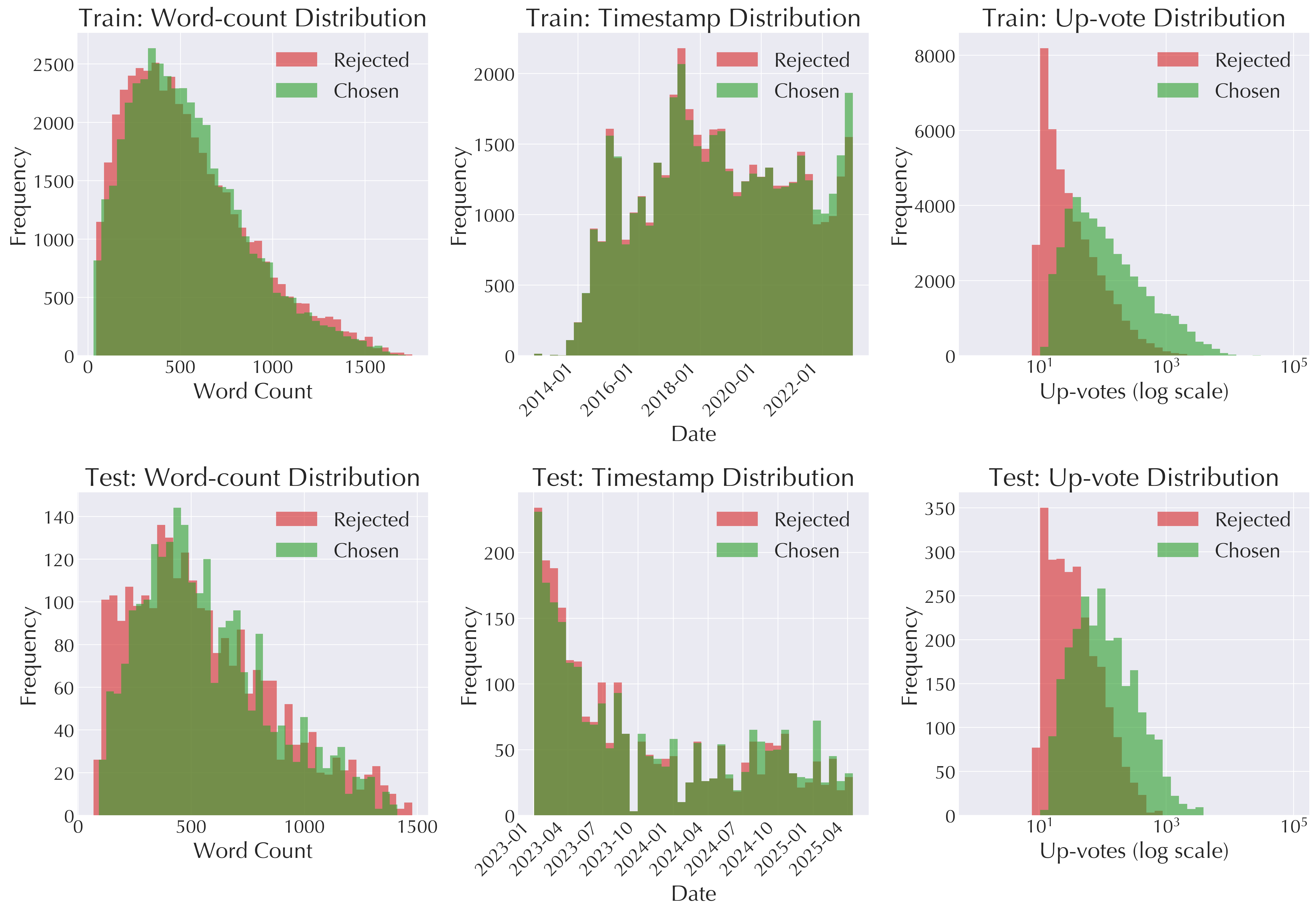}
    \caption{Distributions of word count, date, and upvotes for the LitBench test- and train-set.}
    \label{fig:dataset-final}
\end{figure}

LitBench consists of 2,480 pairwise comparisons composed of 3,543 total stories. These stories have an average length of 550 words, and story length is right-skewed, with a tail of longer stories. The data is exclusively sourced from after January 2nd, 2023. This guarantees the data have no overlap with our training dataset, as well as enabling true zero-shot evaluation of some current language models with earlier training cut-offs. We find that many of our rejected stories have an upvote-count near our prescribed minimum of 10 upvotes, with a long tail of higher rated stories. Our chosen stories have a minimum upvote count of 14, due to our decision to prune pairwise examples with an upvote differential of less than 25\% of the chosen response. These distributions are shown in \ref{fig:dataset-final}

The training dataset consists of 50,309 unique stories that are used in 43,827 pairwise-examples. The distribution is similar to the test set with respect to story lengths and upvote distribution, but the stories in the training set come strictly before 2023. The vast majority of stories were posted between 2014 and 2022. We confirm the quality of annotated training set is indeed higher than the original set by comparing reward models trained on them in Section \ref{sec:results}.

\subsection{Qualitative Analysis of ``Chosen'' Samples}
What is the character of ``winning'' writing samples? To determine this qualitatively, we read and annotated 50 pairwise writing samples from LitBench. We present a few observations below.

\textbf{Why do stories win?} The preferred stories often contain an unexpected twist or surprising humor; we observed many clever punchlines and wordplay. For example, we read about a tyrant queen who won over her opposition not by warfare but absurdist politeness, subverting reader expectations. Another told the story of a woman and her powerful captor named ``Decimator''. The story played with dark themes, and its humor toed the line between edgy and obscene, amusing us (and Reddit users too!)

\textbf{Why do stories lose?} Although some stories were difficult to distinguish, many felt dry and lacked emotional qualities. We found some stories were challenging to finish, due to confusing narratives or strange diction. We were perplexed by a science-fiction tale with too many characters: there was an ``era-model'' soldier with a ``chip'', a woman named ``Gabby'', a shape-shifting monster, and more -- too many characters for a short story; not to mention, the reader was faced with a rapidly shifting point of view. Of note, grammatical errors and narrative incoherence, while present in occasional losing samples, \textit{do not} generally characterize them.


\section{Training and Evaluation Protocols}
We evaluate various approaches to verification including zero-shot Bradley-Terry discriminative reward models, and generative reward models with and without chain-of-thought generation. \citep{wei2022chain}.
\paragraph{Bradley-Terry Discriminative Reward Models} We train a discriminative reward model using the Bradley–Terry (BT) formulation~\citep{bradley1952rank}, where each writing sample in a pair is scored independently, and the model is trained to assign higher reward to the preferred sample. Given reward scores $r_{\text{chosen}}$ and $r_{\text{rejected}}$, the loss is defined as:
\[
\mathcal{L}_{\text{BT}} = -\log \sigma(r_{\text{chosen}} - r_{\text{rejected}}),
\]
encouraging separation between better and worse samples. We append a linear layer to the base-model's last hidden state and then fine-tune all weights of the combined regression model. Accuracy is calculated as the percentage of cases where $r_{\text{chosen}} > r_{\text{rejected}}$.

\paragraph{Generative Reward Models}
Generative reward models have been shown to perform well in math and coding domains, particularly for out of distribution data.(\citep{mahan2024generative};\citep{zhang2024generative}). Generative verifiers treat classification as autoregressive generation by conducting supervised finetuning with cross-entropy loss on the predictions of an instruction-tuned model. Chain-of-thought (CoT) can also be incorporated into this process by finetuning chains of thought that precede and describe the prediction that follows. Here, we train two versions of generative reward models (GenRM): (1) GenRM - to predict which single token between ``A'' and ``B'' is going to be selected, (2) GenRM-CoT - to reason before selecting a preferred story distilled using GPT4.1 generated rationales. At test time, we randomly shuffle the chosen and rejected stories between option A and B to avoid position bias, GenRM's verdicts were collected at temperature=0 with one sample.

\paragraph{Zero-shot LLM Judges} Off-the-shelf, LLM judges are presented unlabeled stories A and B, and asked for a verdict indicating their preference (e.g. ``A'' or ``B'') between the pairwise samples. In particular, we instructed judges to form \textit{explanations} prior to verdict generation. To account for the known position bias in LLM judges \citep{ye2024justiceprejudicequantifyingbiases}, we take the average performance of two sets of pairs, permuting the position of the stories. We selected the LLM-as-judge template prompt, specifying evaluation criteria and output format, by selecting the template with the highest precision in a validation set sampled from the training set among five hand-constructed prompts. We chose not to use automatic prompt optimization tools, such as TextGrad (\citep{yuksekgonul2025optimizing}), as we observed that these methods resulted in poorer prompt performance. For further discussion on prompt optimization, prompt templates and response structure, see \ref{sec:appendix}).  We apply the judge methodology to a selection of state-of-the-art proprietary and open-source LLMs, and these results are demonstrated as baselines in Figure \ref{fig:model_scaling_laws} and \ref{fig:model_comparison}.

\section{Results and analysis}
\label{sec:results}
This work is motivated by the premise that verification in creative, open-ended domains can be operationalized with learned reward models. We offer LitBench pursuant with this motivation, and defend its utility by:
\begin{itemize}
    \item \textit{Validating} the construction of the dataset by benchmarking trained reward models, and evidencing reward model generality with online studies.
    \item \textit{Characterizing} LLM-based methods to verify human writing, by comparing cross-model performance and analyzing their reasoning text.
\end{itemize}

\begin{figure}[ht]
  \centering
  \includegraphics[width=0.95\linewidth]{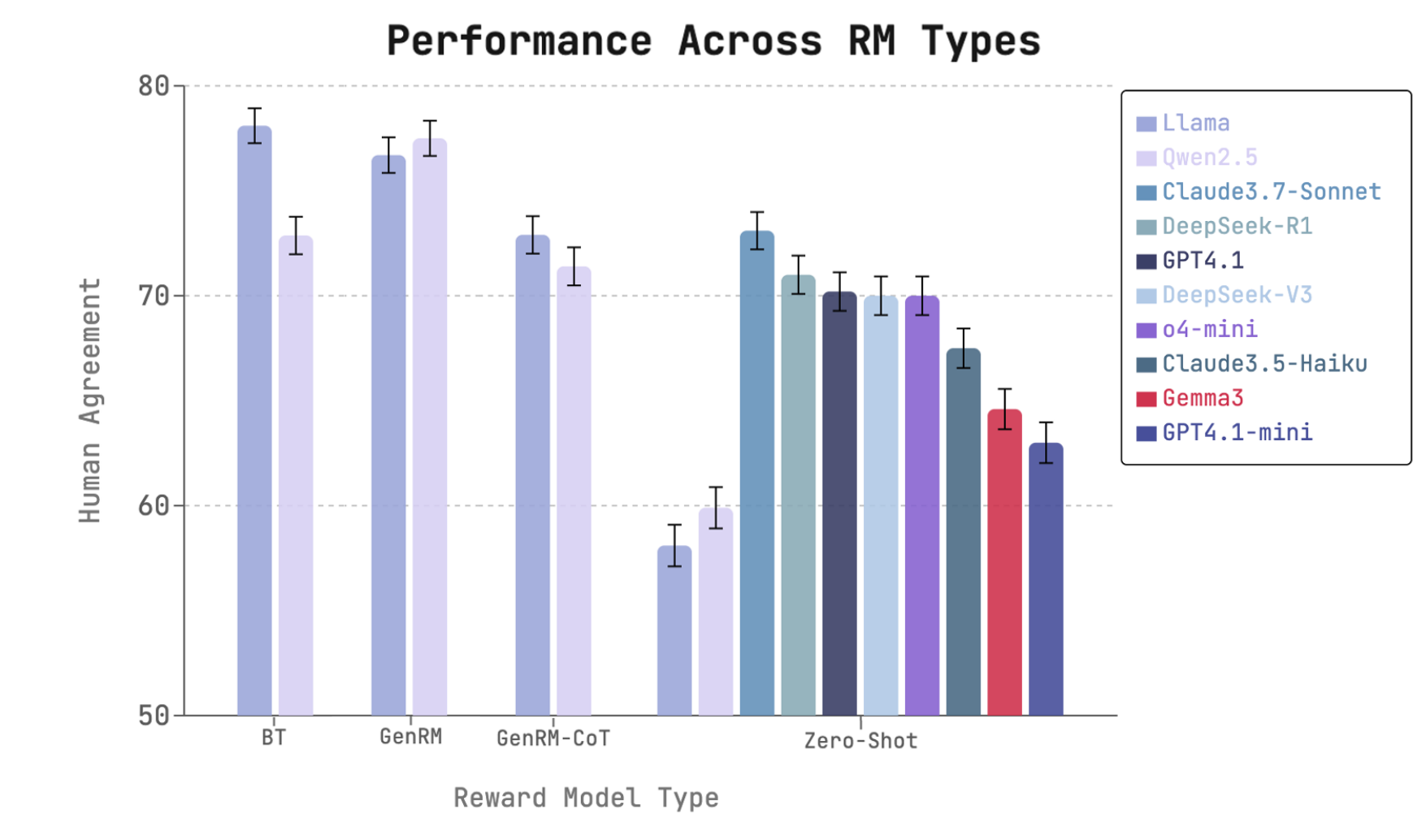}
  \caption{Trained verifiers outperform zero-shot LLM-judges on LitBench. Claude3.7-Sonnet is the strongest zero-shot model. BT verifiers are competitive with GenRMs, but GenRMs with CoTs perform worse. The sizes of Qwen, Llama and Gemma backbones are 7B, 8B and 12B, respectively.}
  \label{fig:model_comparison}
\end{figure}

\paragraph{BT and Generative Reward Models Outperform Zero-Shot LLMs}
We offer comparative reward model performance in Figure \ref{fig:model_comparison}. The best Bradley–Terry reward model (Llama-8B) fine‑tuned on LitBench training set achieves 78\% human agreement, marginally surpassing the best generative verifier (GenRM-Qwen). Both GenRM and BT reward models significantly outperform the strongest zero‑shot judge (Claude-3.7-Sonnet, 73\%). Interestingly, adding chain‑of‑thought reasoning to GenRMs lowers accuracy to 72\%, indicating that explicit sequential reasoning, while beneficial in math and coding tasks, introduces textual noise when judging narrative quality. Zero‑shot performance scales unpredictably with backbone size; SoTA OpenAI, Anthropic and Deepseek models sit in the 70\% range, while smaller open‑source models hover near chance‑plus (56–60\%). These results underscore that targeted preference fine‑tuning dominates parameter count for creative‑writing evaluation, and that discriminative objectives remain the most reliable choice in this domain.

\begin{figure}
    \centering
    \includegraphics[width=1\linewidth]{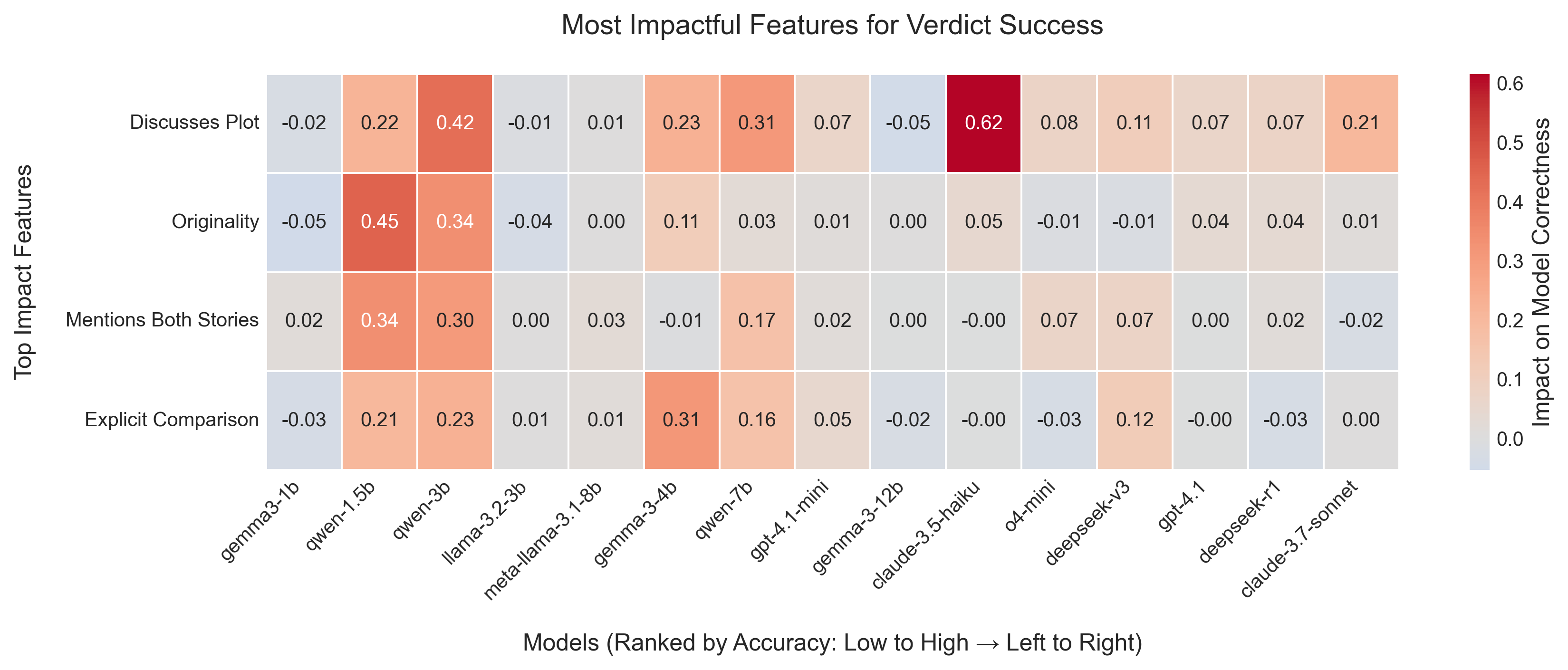}
    \caption{Qualities of explanation text that impact verdict accuracy.}
    \label{fig:topic-cot}
\end{figure}

\paragraph{Reasoning Degrades Verdict Accuracy}
Despite the general success of CoT-based performance bootstrapping (as discussed in Section \ref{sec:verification}), here CoTs actually degraded reward model performance. To examine this, we computed statistics on explanation text produced by judge models, and correlated these features with verdict accuracy. We present characteristics, inspired by creative writing pedagogy \citep{sellers2021practicecreativewriting}, most predictive of verdict accuracy in Figure \ref{fig:topic-cot}. Among all models, discussions of the \textit{plot} are most predictive of correctness (though particularly for Anthropic models) correlated with a +14.8\% higher correctness among all models. However, most of the explanation text features had minimal relation with subsequent verdict accuracy.

\begin{figure}[ht] 
  \centering  
  \includegraphics[width=1.0\linewidth]{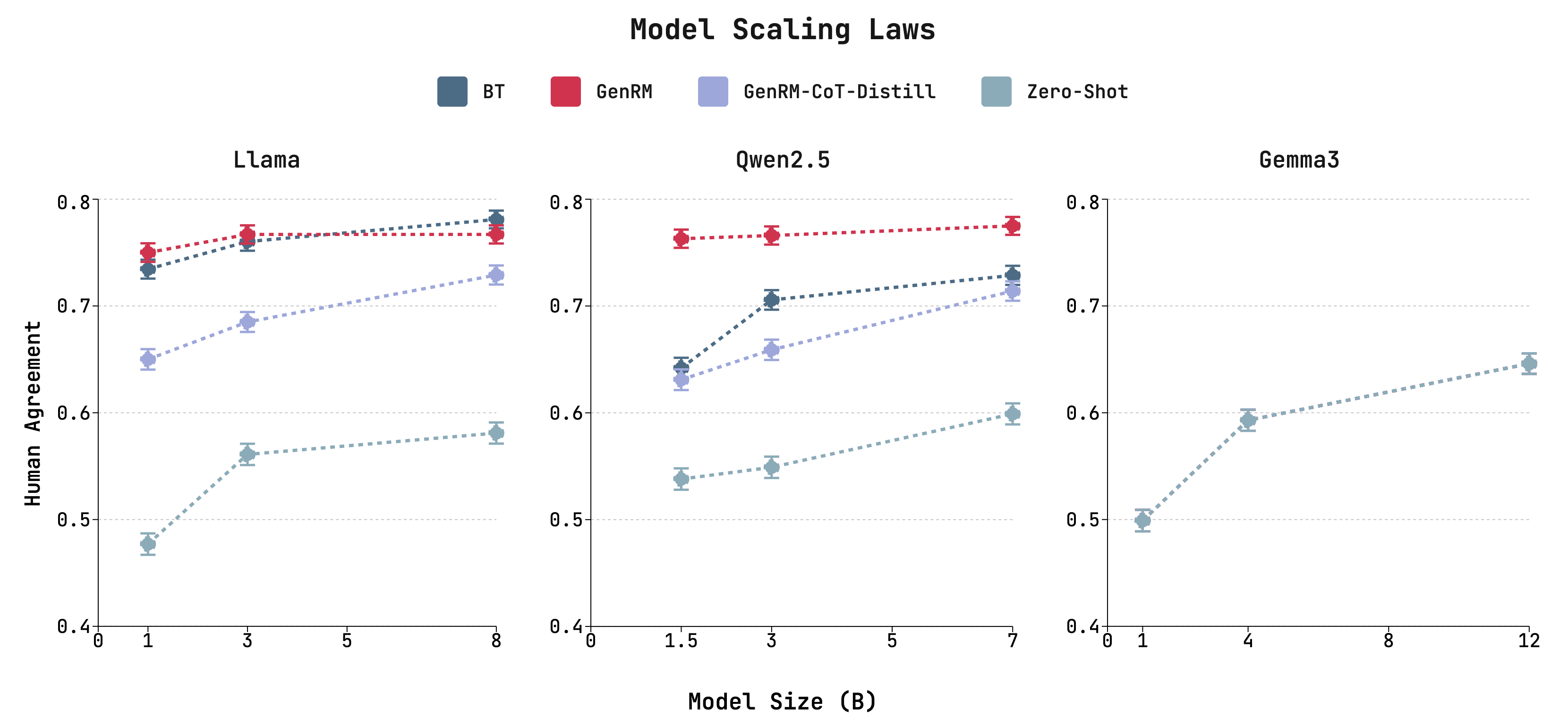}
  \caption{Human agreement scaling is inconsistent with model size for different types of RMs.}
  \label{fig:model_scaling_laws}
\end{figure}

\paragraph{Performance Scales Differentially by Reward Model}
 Figure \ref{fig:model_scaling_laws} shows the performance improves at different magnitudes as the model size increases across all types of reward models. GemRM-CoT starts at lower performance at lower model sizes for both Llama and Qwen backbone, but steadily improves to 74\%. However, GenRM without CoT have no significant improvement across model sizes, suggesting a much smaller model (1B or 1.5B) can be used to obtain similar performance. The performance of Bradley-Terry models has a notable performance difference due to the varied backbone, particularly in smaller models (1B/1.5B and 3B). For zero-shot judges, we observe similar effect that performance improves meaningfully as the size increase.

 \begin{figure}[ht]
  \centering
  \includegraphics[width=0.95\linewidth]{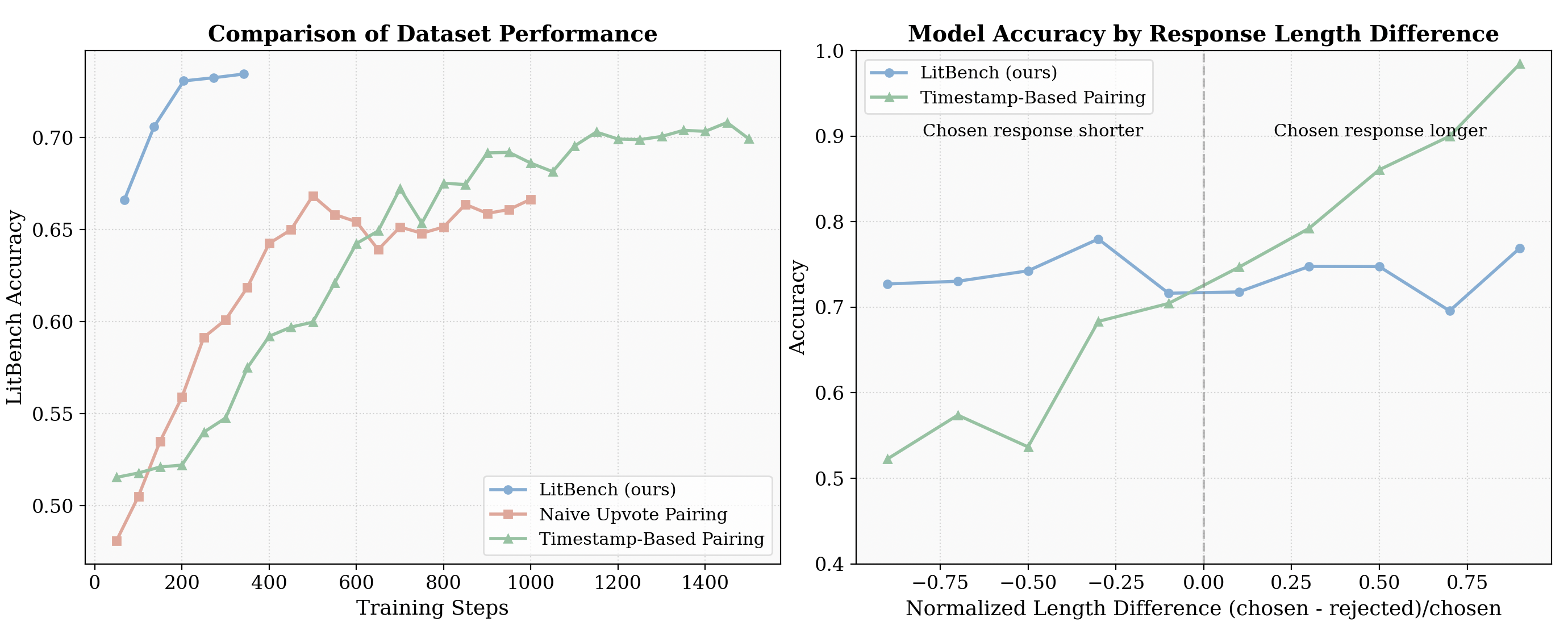}
  \caption{Naive upvote pairing and naive timestamp and upvote pairing saturate at lower accuracy than the LitBench training set. Naive timestamp and upvote pairing alone produces a length biased verifier. All models are BT RMs fine-tuned on a Llama-1B backbone.}
  \label{fig:dataset-ablations}
\end{figure}

\paragraph{Validating Data Filtration Methodology} We further confirm our curation process by training ablative BT reward models on datasets produced by different filtering strategies and evaluating on the debiased LitBench test-set. We create a lightly filtered version of the original training set that only removes pairs containing stories that have less than 10 upvotes and pairs based on upvote difference, resulting in 395k pairs. We also create an unfiltered version paired by timestamp and upvote difference, resulting in 1.03M pairs. We train BT reward models with a Llama-3.2-1B backbone on these datasets. Despite having significantly more examples, we find that performance on LitBench without pairing by timestamp saturates at much lower levels (65\%). Without our length filtering, we find saturation at 70\%, but we also find that the the reward model is length-biased, strongly preferring the longer of the two stories in most cases. These results of this experiment are shown in Figure \ref{fig:dataset-ablations}.

\begin{figure}[ht]
  \centering
  \includegraphics[height=0.4\linewidth]{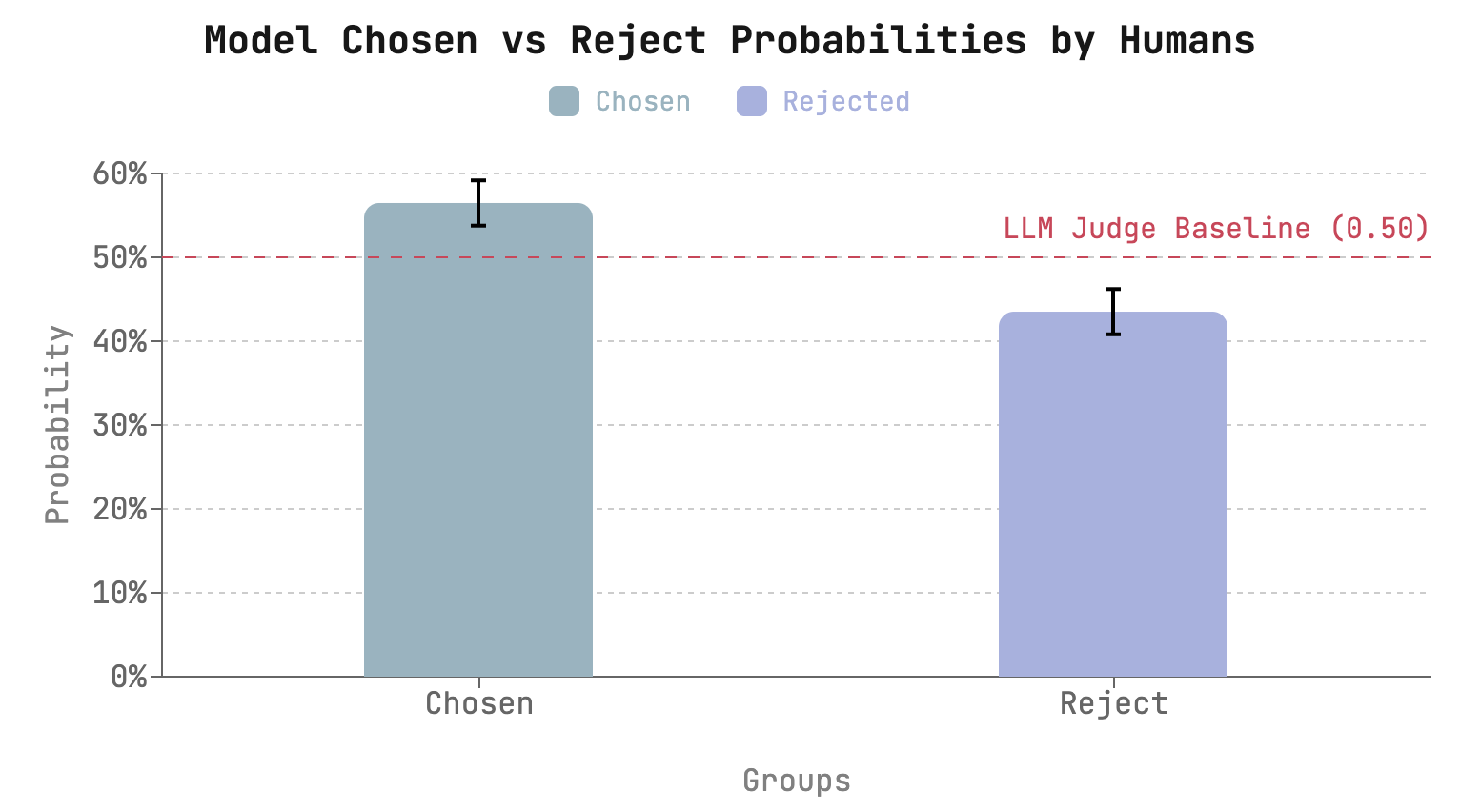}
  \caption{Human preference alignment with reward model on generated writing pairs.}
  \label{fig:human-eval}
\end{figure}

\paragraph{Human Experiments} We generate 64 stories each seeded from 40 LitBench prompts using GPT 4.1 and GPT 4o, and then rank them with our Llama-8B-based Bradley-Terry reward model. In an online human studies with 46 U.S./U.K. crowd‑workers (10-13 annotators per pair), we evaluate human agreement with the RM-determined best and worst stories for each prompt. Figure \ref{fig:human-eval} indicates that annotators selected the RM‑preferred story 57\% of the time versus 41\% for the rejected story, surpassing the best LLM judge (Claude-3.7-Sonnet) which performs at chance. These results confirm that preference fine‑tuning on Reddit labels generalizes to fresh creative‑writing prompts, yet the 40\% disagreement rate underscores substantial head‑room for richer supervision signals—such as rubric‑based feedback or rationale distillation—to further align automatic rewards with human literary taste.

\section{Discussion}
This work shows that the strongest off-the-shelf LLM-judges begin approaching the performance of fine-tuned, domain-specific models in creative writing evaluation. Thus, in the absence of costly human preference data, proprietary LLM-judges appear to be a viable substitute for trained verifiers. When training data is available, GenRMs are most consistently accurate across across different base models and sizes. Our results also call into question the use of GenRMs trained on chains of thought. Lastly, our human evaluation reveals that performance on LitBench generalizes to evaluation of newly LLM-generated stories, suggesting that future work might make use of a strong verifier to improve latent creative writing generation capabilities.

\section{Limitations}
\label{sec:limitations}
A key limitation of our work stems from the assumption, inherited from \citep{ethayarajh2023stanford}, that upvotes on Reddit contain information about human preferences. Though we experimentally validate our dataset via human evaluation, it is unclear what other underlying factors may be encoded in upvote information. For example, \citep{kassaeyan2016factors} report that the decision to upvote a post on social networks is at least partially driven by personal and social mechanisms, including individuation, perceived behavioral control, and altruism. 

Our extension of human preferences to determine writing quality is further complicated by the question of subjectivity in writing evaluation. There is a body of work showing how measured features of writing can correlate with human ratings in aggregate \citep{zedelius2019beyond, mcnamara2010linguistic}. Moreover, authorship on fair writing evaluation in a classroom setting establishes precedent for objectivity in writing \citep{Weigle2002}. 

\cite{elam2023poetry} argues that writing generated artificially ``renders meaning senseless'' via representing realities and contexts that do not actually occur within history. In a similar way, our verifiers are limited by their removal from legitimate, individual human experiences that ground all creative writing. We acknowledge that no automated quality verifier can fully capture the social value of arbitrary text in real-world contexts, and we regret any implication to the contrary. Our dataset comes from Reddit, which has been reported to demographically skew male, educated, and middle-aged \citep{duarte2025reddit, agrawal2016reddit}. Ultimately, our benchmark and accompanying dataset reflect the consensus preferences of these groups.


\clearpage           

\begingroup
\raggedbottom

\appendix
\label{sec:appendix}
\section{Appendix}

\subsection{LLM-as-judge Raw Results}

\begin{table}[htbp]
  \centering
  \begin{tabular}{@{}l S[table-format=1.3] S[table-format=4.1]@{}}  
    \toprule
    \textbf{Model} & \textbf{$\text{Acc}_{\text{Jan23}}$} & \textbf{Avg.\ Expl.\ Len.} \\
    \midrule
    claude-3-5-haiku        & 0.675 & 292.4 \\
    \textbf{claude-3-7-sonnet} & \textbf{0.731} & 280.2 \\
    \addlinespace
    gpt-4.1                 & 0.702 & 202.3 \\
    gpt-4.1-mini            & 0.630 & 246.7 \\
    o4-mini                 & 0.700 & 131.5 \\
    \addlinespace
    deepseek-v3             & 0.700 & 167.4 \\
    deepseek-r1             & 0.710 & 142.8 \\
    \addlinespace
    gemma-3-12b-it          & 0.657 & 497.0 \\
    \addlinespace
    llama-3.1-8b            & 0.581 & 332.0 \\
    \addlinespace
    qwen-2.5-7b             & 0.599 & 174.0 \\
    \bottomrule
  \end{tabular}
  \caption{LLM-as-a-judge evaluation results by model on LitBench.}
  \label{tab:model_results_families}
\end{table}

\subsection{LLM-as-judge Prompt Optimization}

\textbf{Motivation.}
We sought to give out-of-the-box LLM evaluators the best chance of performing accurately on this benchmark.  
LLM-based judges have been demonstrated to be extremely sensitive to prompt content (CITE).  
In this experimental setting, prompts enable (a) introduction of criteria for judges to utilize in inferring verdicts, and (b) specification of an output format to ensure easily parsed results.


\textbf{Strategy.}
There are numerous prompt-optimization libraries that automatically 'differentiate' the prompt text to improve accuracy on a given evaluation metric.  
However, after some experimentation with these, we opted to apply methods of our own design to optimize the prompts, following the approach below. \\ 
\\
\textbf{Goal:} Select an optimized prompt for each family of models (e.g. Llama).

\textbf{Optimization Method.}
\begin{enumerate}[leftmargin=*,
                itemsep=2pt,      
                parsep=0pt,
                topsep=1pt]       
  \item Hand-construct six 'template' prompts, each introducing different criteria for the judge to use when generating a verdict.
  \item Standardize output format: request JSON objects from large instruction-tuned models; request plaintext from smaller models.
  \item Using a midrange model for each family, evaluate each prompt with a validation set ($n=500$) drawn from the training set to avoid bias.
  \item Adopt the prompt, by family, that yields the highest accuracy.
  \item Append the standardized output format instruction, depending on the model size and capacity to follow instructions.
\end{enumerate}



\subsection{Prompt Templates.} \hspace{0.5em}%


\textbf{1. Writer-ly Criteria}\\

\begin{tcolorbox}[breakable,
                  colback=blue!5!white,
                  colframe=blue!75!black,
                  title=\textbf{Writer-ly Criteria Prompt}]
\begin{Verbatim}[breaklines, breaksymbolleft=, fontsize=\small]
You're evaluating creative writing responses A and B.

Compare them based on these dimensions:
- Imagery: vivid descriptions and sensory details
- Tension: dramatic interest and conflict
- Pattern: structural elements and composition
- Energy: engaging style and dynamic writing
- Insight: meaningful ideas and depth

IMPORTANT: Your answer MUST use EXACTLY this format:
Reasoning: [brief comparison]
Preferred: [A or B] (state which one is better)

Example format:
Reasoning: Response B has stronger imagery and tension.
Preferred: B
\end{Verbatim}
\end{tcolorbox}
\vspace{2em}
\textbf{2. Alternative Criteria}\\

\begin{tcolorbox}[breakable,
                  colback=blue!5!white,
                  colframe=blue!75!black,
                  title=\textbf{Alternative Criteria Prompt}]
\begin{Verbatim}[breaklines, breaksymbolleft=, fontsize=\small]
Evaluate creative writing responses A and B.

Consider these aspects:
- Originality: unique concepts, unexpected elements
- Imagery: sensory language and descriptions
- Emotional impact: how the writing affects the reader
- Coherence: logical flow and narrative structure
- Technical skill: language use and style

FORMAT REQUIRED:
Reasoning: [your evaluation]
Preferred: [A or B]
\end{Verbatim}
\end{tcolorbox}
\vspace{2em}

\textbf{3. Minimal Instruction}\\

\begin{tcolorbox}[breakable,
                  colback=blue!5!white,
                  colframe=blue!75!black,
                  title=\textbf{Minimal Instruction Prompt}]
\begin{Verbatim}[breaklines, breaksymbolleft=, fontsize=\small]
Compare responses A and B for creative writing quality.
MUST follow this format:
Reasoning: [brief analysis]
Preferred: [A or B]
\end{Verbatim}
\end{tcolorbox}
\vspace{2em}
\textbf{4. Reddit-Minimal}\\

\begin{tcolorbox}[breakable,
                  colback=blue!5!white,
                  colframe=blue!75!black,
                  title=\textbf{Reddit Minimal Instruction Prompt}]
\begin{Verbatim}[breaklines, breaksymbolleft=, fontsize=\small]
You are evaluating two creative writing responses (A and B) to the same writing prompt.
Your task is to predict which response would receive more upvotes from the Reddit community.

Your verdict MUST follow this exact format:
Reasoning: [explain which response would likely get more Reddit upvotes and why]
Preferred: [A or B] (the one you predict would get more upvotes)
\end{Verbatim}
\end{tcolorbox}
\vspace{2em}
\textbf{5. Reddit-Verbose} \\

\begin{tcolorbox}[breakable,
                  colback=blue!5!white,
                  colframe=blue!75!black,
                  title=\textbf{Reddit-Verbose Prompt}]
\begin{Verbatim}[breaklines, breaksymbolleft=, fontsize=\small]
You are evaluating two creative writing responses (A and B) to the same writing prompt. These responses are similar to those posted on Reddit writing subreddits like r/WritingPrompts.

Your task is to predict which response would receive more upvotes from the Reddit community. Reddit users typically upvote creative writing that is engaging, original, well-written, and emotionally resonant.

When making your prediction, consider what makes content popular on Reddit:
- Originality and uniqueness of ideas
- Engaging narrative style and pacing
- Emotional impact and relatability
- Clever twists or satisfying conclusions
- Technical quality of writing

This is an experiment to test how well language models can predict human preferences in creative writing as expressed through Reddit's voting system.

Your verdict MUST follow this exact format:
Reasoning: [explain which response would likely get more Reddit upvotes and why]
Preferred: [A or B] (the one you predict would get more upvotes)
\end{Verbatim}
\end{tcolorbox}
\vspace{2em}

\textbf{6. Reddit-Verbose Permuted}\\

\begin{tcolorbox}[breakable,
                  colback=blue!5!white,
                  colframe=blue!75!black,
                    title=\textbf{Reddit-Verbose Permuted}]
\begin{Verbatim}[breaklines, breaksymbolleft=, fontsize=\small]
You are tasked with evaluating two creative writing responses (A and B) to the same prompt. Your goal is to predict which response would garner more upvotes from the Reddit community, specifically in writing subreddits like r/WritingPrompts. 

Consider the following key dimensions for your evaluation:
- Creativity and Originality: How unique are the ideas presented?
- Narrative Engagement: Is the storytelling captivating and immersive?
- Emotional Resonance: Does the piece evoke feelings or relatable experiences?
- Surprise and Satisfaction: Are there clever twists or fulfilling conclusions?
- Writing Quality: Is the grammar, style, and structure polished?

Your output must strictly follow this format:
1. Reasoning: [Explain which response is likely to receive more Reddit upvotes, citing specific strengths and weaknesses.]
2. Preferred: [A or B]

Be concise and clear in your assessment, adhering to the format above.
\end{Verbatim}
\end{tcolorbox}


\section{Dataset Licenses and Access}
\label{sec:data}
All data is publicly accessible via \href{https://huggingface.co/collections/SAA-Lab/litbench-68267b5da3aafe58f9e43461}{the SAA-Lab/LitBench collection on Hugging Face}. Code to use this dataset is available on \href{https://github.com/drfein/LitBench/tree/main}{GitHub}.

Our train set content is sourced from \href{https://huggingface.co/datasets/euclaise/WritingPrompts_preferences}{euclaise/WritingPrompts\_preferences} on Hugging Face, which has an MIT-license. In our test set, we release ids of ~3.5k Reddit comments from \texttt{r/WritingPrompts}, along with code to rehydrate from the reddit api. We acknowledge that Reddit users retain copyright over their individual comments, and we do not claim ownership or offer any re-licensing of this content. We contacted Reddit in advance of this release to clarify acceptable use under their API terms. As of submission time, we have not received a response.

\section{Compute Usage}
\label{sec:compute}
All training runs and evaluation was done on our internal cluster using a node with 128 CPU cores, 8 NVIDIA A40 GPUs each with 48GB of VRAM, and a total of 732GB of system RAM. Training verifiers took between 3 hours for 1B-parameter models and up to one day for 8B parameter models. The total compute used, including failed runs, data ablations, and generation for LLM-as-a-judge is estimated at 500 GPU-hours on NVIDIA A40.

\section{Training Hyperparameters}
\label{sec:hps}
For all training runs, we use an effective batch size of 128 examples, a learning rate of 1e-5 with a warmup ratio of 10\%. We train in \texttt{bfloat16} and use AdamW as our optimizer.

\bibliographystyle{plainnat}
\bibliography{references}

\end{document}